\documentclass[letterpaper, 10 pt, conference]{ieeeconf}  

\IEEEoverridecommandlockouts                              
\usepackage[utf8]{inputenc}
\usepackage[T1]{fontenc}
\usepackage{amsmath}
\usepackage{graphicx}
\usepackage{multirow}
\usepackage{balance}

\usepackage{titlesec}
\renewcommand{\thesection}{\arabic{section}}

\titleformat{\section}
  {\normalfont\bfseries\centering} 
  {\thesection}                    
  {1em}                            
  {\MakeUppercase}                

\title{\LARGE \bf Physics-Informed Image Restoration via Progressive PDE Integration}

\author{Shamika Likhite*, Santiago López-Tapia, Aggelos K. Katsaggelos
\thanks{* denotes corresponding author}
\\ Department of Electrical and Computer Engineering\\
Northwestern University, Evanston, IL, USA
\\shamikalikhite2024@u.northwestern.edu, \{santiago.lopeztapia, a-katsaggelos\}@northwestern.edu
}


\begin{document}

\maketitle
\thispagestyle{empty}
\pagestyle{empty}

\begin{abstract}
Motion blur, caused by relative movement between camera and scene during exposure, significantly degrades image quality and impairs downstream computer vision tasks such as object detection, tracking, and recognition in dynamic environments. While deep learning-based motion deblurring methods have achieved remarkable progress, existing approaches face fundamental challenges in capturing the long-range spatial dependencies inherent in motion blur patterns. Traditional convolutional methods rely on limited receptive fields and require extremely deep networks to model global spatial relationships, while transformer-based approaches with self-attention mechanisms can model global context but introduce additional architectural complexity. These limitations motivate the need for alternative approaches that incorporate physical priors to guide feature evolution during restoration. In this paper, we propose a novel motion deblurring framework that integrates physics-informed partial differential equation (PDE) dynamics into state-of-the-art restoration architectures through a progressive training strategy. By leveraging advection-diffusion equations to model feature evolution, our approach naturally captures the directional flow characteristics of motion blur while enabling principled global spatial modeling. The advection component directs information flow along image structures to preserve edges and textures, while the diffusion component provides adaptive smoothing to reduce artifacts during the restoration process. Our PDE-enhanced deblurring models achieve superior restoration quality with minimal overhead, adding only approximately 1\% to inference GMACs (0.6 to 2.4 GMACs depending on the base architecture) while providing consistent improvements in perceptual quality across multiple state-of-the-art architectures. Comprehensive experiments on standard motion deblurring datasets (GoPro, RealBlur-R, and RealBlur-J) demonstrate that our physics-informed approach improves PSNR and SSIM significantly across four diverse architectures, including FFTformer, NAFNet, Restormer, and Stripformer. These results validate that incorporating mathematical physics principles through PDE-based global layers can enhance deep learning-based image restoration, establishing a promising direction for physics-informed neural network design in computer vision applications.
\end{abstract}


\section{INTRODUCTION}

Motion blur significantly degrades image quality and reduces accuracy in subsequent computer vision tasks such as object classification, scene understanding, and activity recognition. For instance, analyzing high-speed photography for motion capture or automated analysis is crucial in sports analytics and surveillance applications. Such dynamic imaging scenarios inherently involve rapid subject movements and camera operations, introducing substantial motion blur that severely compromises the accuracy of object detection, feature extraction, and recognition systems. Similarly, mobile photography and handheld camera systems frequently encounter motion blur due to camera shake, subject movement, or inadequate stabilization, resulting in degraded image quality that significantly impacts downstream computer vision applications in real-world scenarios.

A fundamental challenge in image deblurring is effectively capturing long-range spatial dependencies in motion blur patterns. Traditional convolutional approaches rely on limited receptive fields and require very deep networks to model global spatial relationships, while recent transformer-based methods achieve global modeling through self-attention mechanisms but with additional architectural complexity. Current state-of-the-art image restoration methods typically employ data-driven optimization without explicitly incorporating physical priors about the blur formation and restoration process. This motivates exploring physics-informed approaches that can guide feature evolution through mathematical constraints while maintaining restoration quality.

Inspired by the work "Condensing CNNs with Partial Differential Equations,"~\cite{kag2022}, we propose a novel methodology that incorporates PDE-informed principles into image restoration networks. Unlike traditional approaches that rely solely on data-driven optimization, our method leverages the mathematical foundations of differential equations to model feature evolution during the restoration process. Specifically, we introduce PDE-based global layers that capture the directional flow characteristics inherent in motion blur patterns through advection-diffusion dynamics. To address the gradient stability challenges that arise when integrating iterative PDE solvers into deep networks, we develop a progressive training strategy that gradually increases the number of PDE iterations while correspondingly reducing the temporal step size to maintain constant total integration time. This progressive approach enables stable convergence while allowing the network to benefit from finer temporal discretization as training progresses. By incorporating mathematical priors that model blur formation as a physical process, we achieve improved restoration quality across multiple state-of-the-art architectures with minimal additional computational overhead.

In this paper, our main contributions are:
\begin{enumerate}
    \item We propose the incorporation of a Neural Network layer inspired by the advection-diffusion PDE into deep learning image restoration models through a progressive training strategy. This allows feature evolution to be guided by advection–diffusion dynamics, enabling more effective restoration while preserving global spatial interactions. 
    \item  We show that incorporating our training method into existing architectures leads to substantial improvements in motion deblurring, enhancing perceptual quality and downstream vision task performance without significantly increasing computational complexity.
\end{enumerate}

\begin{figure*}[t]
    \centering
    \includegraphics[width=\textwidth]{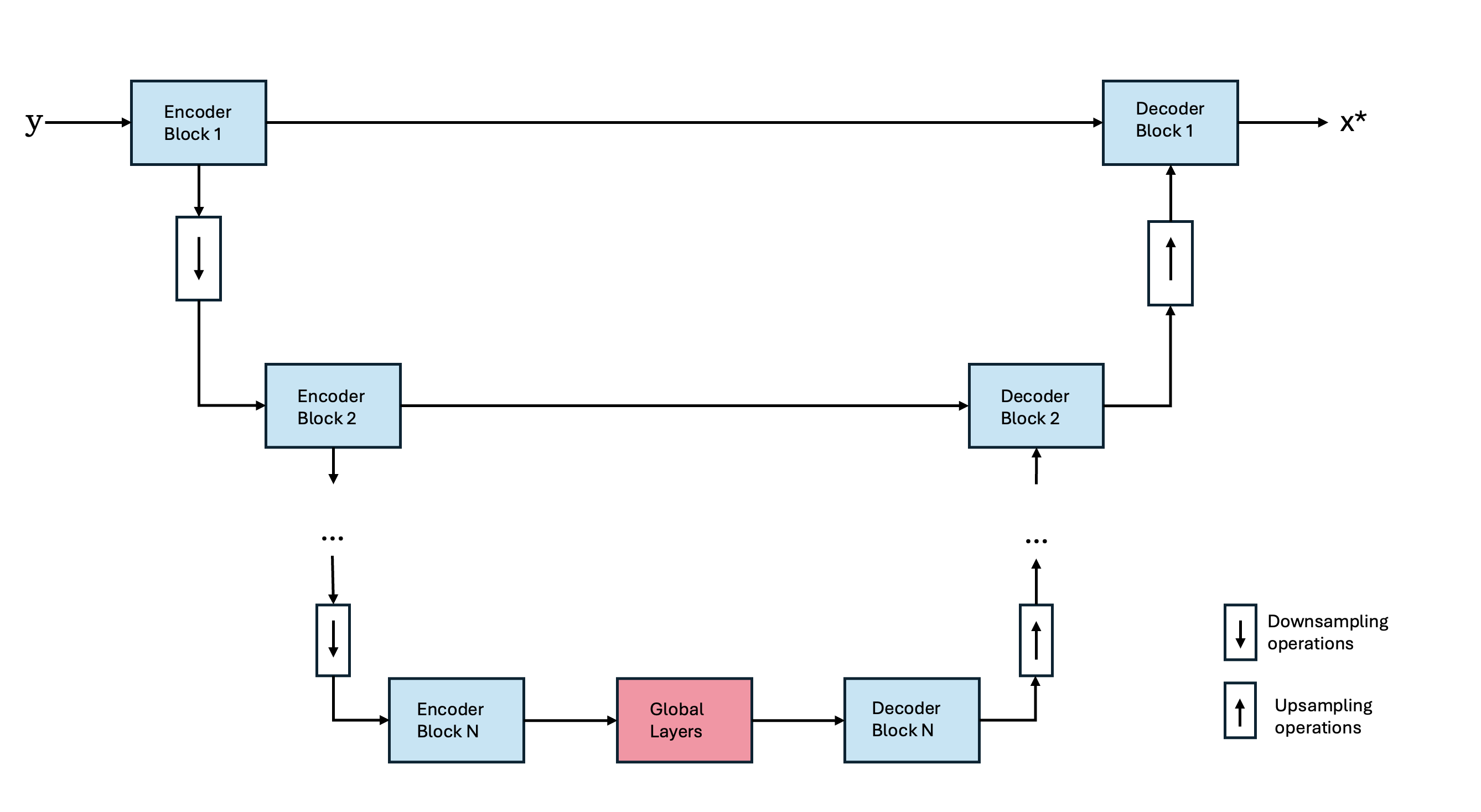}
    \caption{High-level architecture of the image restoration network. The network follows a U-Net style encoder-decoder structure, where the input degraded image y is progressively downsampled through encoder blocks (Encoder Block 1 to N) via max pooling operations. At the bottleneck, global layers process the latent representation at the coarsest scale. The decoder path progressively upsamples the features through decoder blocks (Decoder Block N to 1) using upward convolutions to reconstruct the restored image x*. Skip connections between corresponding encoder and decoder blocks preserve spatial information across scales. The global layers at the bottleneck serve as the primary location for integrating physics-informed processing. }
\label{fig:architecture}
    \label{fig:unet_pde}
\end{figure*}
\section{Related Work}

\subsection{Evolution of Image Deblurring Methods}
Image deblurring has evolved significantly from early traditional approaches to sophisticated deep learning methods that leverage advanced architectural designs. Traditional deconvolution-based methods [1,2] rely on mathematical models of the blur process, often assuming uniform blur kernels and struggling with complex, spatially-varying motion patterns. These approaches suffer from computational inefficiency and suboptimal restoration quality due to their reliance on hand-crafted priors and simplified blur models.
Modern image deblurring methods can be broadly categorized into single-image and multi-scale approaches. Single-image methods [3,4] process individual blurred images directly, focusing on learning effective representations for blur removal. Multi-scale methods [5,6] exploit hierarchical feature representations at different resolutions, enabling better handling of large blur kernels and complex motion patterns but often at the cost of increased computational overhead.
The advent of deep learning has revolutionized image restoration tasks. Convolutional Neural Networks (CNNs) have been extensively employed for image deblurring, with architectures ranging from simple encoder-decoder networks to complex multi-scale designs [7,8]. Recent works have explored attention mechanisms [9] and transformer architectures [10] to better model long-range spatial dependencies crucial for effective blur removal.
NAFNet [11] introduced a novel approach by simplifying network architectures while maintaining competitive performance, demonstrating that carefully designed simple blocks can be highly effective for image restoration tasks. This work highlighted the importance of architectural efficiency and inspired subsequent research into lightweight yet powerful restoration models, reinforcing the principle that architectural simplicity can coexist with high performance in image deblurring applications.
\subsection{Advanced Architectures and Spatial Modeling}
Effective spatial feature modeling remains a fundamental challenge in image restoration. Traditional CNN-based methods [13,14] rely on local convolutions with limited receptive fields, requiring deep networks to capture long-range spatial dependencies necessary for handling large blur kernels. Dilated convolutions [15,16] offer an alternative by expanding receptive fields without increasing computational cost, but they can introduce artifacts and struggle with irregular blur patterns.
Vision transformers have gained significant attention for their ability to model long-range dependencies effectively [22,23]. In image processing, transformers have been applied to capture global spatial relationships, often outperforming CNN-based approaches on various restoration tasks [24,25]. However, the quadratic complexity of self-attention mechanisms poses challenges for high-resolution image processing. Recent works have explored efficient transformer variants and hybrid CNN-transformer architectures to mitigate computational overhead while preserving the modeling capabilities of attention mechanisms [26,27].
\subsection{Mathematical Foundations and Training Methodologies}
The integration of Partial Differential Equations (PDEs) with neural networks has emerged as a promising research direction. The work "Condensing CNNs with Partial Differential Equations"~\cite{kag2022} demonstrated how PDE principles can be incorporated into CNN architectures to achieve better parameter efficiency and improved performance in classification tasks. This approach leverages the mathematical foundations of differential equations to guide feature evolution, resulting in more principled network designs. PDE-informed methods have shown particular promise in capturing continuous spatial relationships, making them well-suited for image processing tasks where understanding spatial dependencies is crucial. However, their application to image restoration problems remains largely unexplored.

While various efficient architectures have been proposed for image restoration~\cite{sandler2017,tan2019}, there is growing interest in approaches that enhance restoration quality through physics-informed priors rather than solely focusing on architectural efficiency. Methods that incorporate mathematical constraints can potentially improve performance by leveraging domain knowledge about the degradation and restoration processes. The challenge lies in effectively integrating such physics-based components into existing deep learning architectures while maintaining stable training dynamics and restoration quality.

\subsection{Motivation and Challenges}

Despite significant progress in image deblurring, current approaches face challenges in effectively modeling long-range spatial dependencies. Traditional CNN-based methods rely on limited receptive fields and require very deep networks to capture global spatial relationships, while transformer-based approaches achieve global modeling through self-attention mechanisms at the cost of additional architectural complexity. Most existing methods employ purely data-driven optimization without explicitly incorporating physical priors about blur formation and restoration processes.

While Kag et al.~\cite{kag2022} demonstrated that PDE-based global layers can be effectively integrated into CNNs for classification tasks, their approach has not been explored for image restoration problems. Image restoration presents unique challenges compared to classification, including spatially varying degradations, the need to preserve fine-grained details and textures, perceptual quality requirements that go beyond pixel-level metrics, and the need for global spatial coherence during reconstruction. More importantly, we have determined experimentally that using PDE steps inside a restoration network brings new stability issues. We hypothesize that, since the solver runs many steps in a row, backpropagation must pass through a long chain, which can make gradient calculation unstable, similar to recurrent neural networks.

\section{Method}
Our approach leverages Partial Differential Equations (PDEs) in a Global Feature Layer \cite{kag2022} that achieves efficient global spatial modeling for image deblurring. Instead of relying on deep networks or large kernels to capture long-range dependencies, we embed an advection-diffusion PDE solver that processes features through lightweight iterative operations. 
As demonstrated in \cite{kag2022}, the use of these layers enables global receptive fields with linear complexity O(K×H×W), with K being the number of iterations of the discretized PDE. This reduces computational overhead compared to traditional CNN approaches(where complexity increases with the size of the receptive field) while maintaining superior restoration quality through physics-informed feature processing.


\subsection{Global Feature Layer and PDE Formulation for Image Restoration}
The Global Feature Layer implements a physics-inspired approach to image restoration where the discretized PDE is used to construct the layer, constraining the output feature map
$H$ following the framework established in [19]. To guide the training of our restoration network, the underlying dynamics are modeled using the advection-diffusion equation:
\begin{equation}
\frac{\partial H}{\partial t} \;=\; \nabla \cdot ( D \nabla H ) \;-\; \nabla \cdot ( \mathbf{v} H ) \;+\; f(I),
\label{eq:advec_diff}
\end{equation}
where $D$ represents the diffusion coefficient(s), $\mathbf{v}$ denotes the velocity field, and $f(I)$ is a source term depending on the input features $I$.

This PDE interprets the restoration process as \textbf{guided particle dynamics} where input pixels evolve under physical constraints:

\begin{itemize}
\item \textbf{Diffusion term} ($\nabla \cdot (D\nabla H)$): Models smoothing where particles move from high-concentration to low-concentration regions. The diffusion coefficient $D$ controls the strength of this spreading. This term handles noise reduction and fills gaps, but without guidance would blur all structures uniformly.

\item \textbf{Advection term} ($\nabla \cdot (\mathbf{v}H)$): Directs information flow along image structures. The velocity field $\mathbf{v} = (u, v)$ determines the direction and speed of particle movement. This allows information to propagate along edges and textures rather than spreading equally in all directions.

\item \textbf{Source term} ($f(I)$): Injects task-specific information from the degraded input $I$, anchoring the solution to observed data and guiding the restoration process.
\end{itemize}

\textbf{Blur/Deblur Modeling}

The advection and diffusion terms work together to enable structure-aware restoration. While the diffusion term smooths the feature map, the velocity field $\mathbf{v}$ guides how this smoothing occurs by directing flow along image structures. This allows the model to smooth homogeneous regions while maintaining sharpness along edges and textures. In deblurring, the velocity field guides pixel movement along structural boundaries, helping recover sharp details without introducing artifacts or over-smoothing.

\subsubsection{2D Image Domain}
Since images are inherently 2D, we expand the velocity and diffusion terms into spatial components:  
$\mathbf{v} = (u,v)$ and $D = (D_x,D_y)$.  
Following the approach in [19], Eq.(1) expands to:
\begin{align}
&\frac{\partial}{\partial t} H(x,y,t) + \frac{\partial}{\partial x}(u(x,y,t)H(x,y,t)) + \frac{\partial}{\partial y}(v(x,y,t)H(x,y,t))\notag\\
&= \frac{\partial}{\partial x}\left(D_x \frac{\partial}{\partial x} H(x,y,t)\right) + \frac{\partial}{\partial y}\left(D_y \frac{\partial}{\partial y} H(x,y,t)\right) + f(I(x,y))
\label{eq:2D_advec_diff}
\end{align}
This 2D formulation captures how pixel information spreads (diffusion) and propagates (advection) across spatial coordinates $(x,y)$, which is essential for handling blur removal and denoising tasks where spatial relationships are critical.

\subsubsection{Iterative Solver Embedded in Training}
Following the discretization scheme proposed in [19], we adopt a finite-difference approach to make the PDE solver practical for end-to-end training. To solve Eq.~\ref{eq:2D_advec_diff} numerically, we discretize the spatial and temporal domains with step sizes $\delta x$, $\delta y$, and $\delta t$. The iterative update rule for feature map evolution becomes:
\begin{align}
L H^{k+1}_{x,y} &= M H^{k-1}_{x,y} - 2 (u_x + v_y) \, \delta t \, H^k_{x,y} + 2 \delta t \, f\!\big(I(x,y)\big) \nonumber \\
&\quad + \, (-A_x + 2B_x) H^k_{x+1,y} + (A_x + 2B_x) H^k_{x-1,y} \nonumber \\
&\quad + \, (-A_y + 2B_y) H^k_{x,y+1} + (A_y + 2B_y) H^k_{x,y-1},
\label{eq:discrete_PDE}
\end{align}
where the coefficients are defined following [19] as:
\begin{align}
L   &= 1 + 2B_x + 2B_y, 
&\quad M   &= 1 - 2B_x - 2B_y, \notag\\
u_x &= \frac{u_{x+1,y} - u_{x-1,y}}{2\delta x}, 
&\quad v_y &= \frac{v_{x,y+1} - v_{x,y-1}}{2\delta y}, \notag\\
A_x &= \frac{u \, \delta t}{2\delta x}, 
&\quad A_y &= \frac{v \, \delta t}{2\delta y}, \notag\\
B_x &= \frac{D_x \, \delta t}{\delta x^2}, 
&\quad B_y &= \frac{D_y \, \delta t}{\delta y^2}.
\end{align}

This discretized solver acts as differentiable layers within our restoration network, ensuring that learned features evolve under PDE-guided dynamics. By adapting the methodology from \cite{kag2022} to image restoration, we enable the training procedure to jointly leverage physics-based evolution and data-driven learning, improving the robustness to unseen blurs, including blur and noise. The iterative nature allows for gradual refinement of features through multiple PDE steps, where each iteration applies the physical constraints while preserving gradient flow for end-to-end optimization. When integrated into existing SOTA restoration architectures, this PDE-based global layer improves network performance. This allows us to reduce the complexity of existing NN-based restoration models while keeping the perceptual quality of the image. Specifically, the global layer's ability to propagate information across spatial coordinates through Eq. 3 reduces the dependency on deep feature extraction hierarchies.

\subsection{Training Stability and Iteration Constraints}

While the PDE-based global layer provides theoretical advantages for spatial information propagation, practical implementation reveals critical stability issues that must be carefully addressed. Our empirical analysis demonstrates that attempting to train directly with multiple PDE iterations ($K > 1$) from the start leads to severe gradient instability. However, this limitation can be overcome through a progressive training strategy that gradually increases iteration count while the network stabilizes.

\subsubsection{Gradient Instability in Multi-Iteration Schemes}

When attempting to train directly with multiple PDE iterations ($K > 1$), we observe severe gradient instability that prevents effective backpropagation. The iterative nature of Eq.~(2) creates a computational graph where gradients must traverse through $K$ sequential finite-difference operations, each involving spatial derivatives and coefficient updates. Each iteration involves multiple multiplications with the learned velocity and diffusion parameters, and spatial derivative approximations. Backpropagating gradients through this deep computational chain compounds these operations, leading to unstable gradient magnitudes that prevent effective training. This is particularly problematic because the diffusion terms $B_x$ and $B_y$ in Eq.~(4) act as smoothing operators.

\begin{figure}[t]
    \centering
    \includegraphics[width=0.5\textwidth]{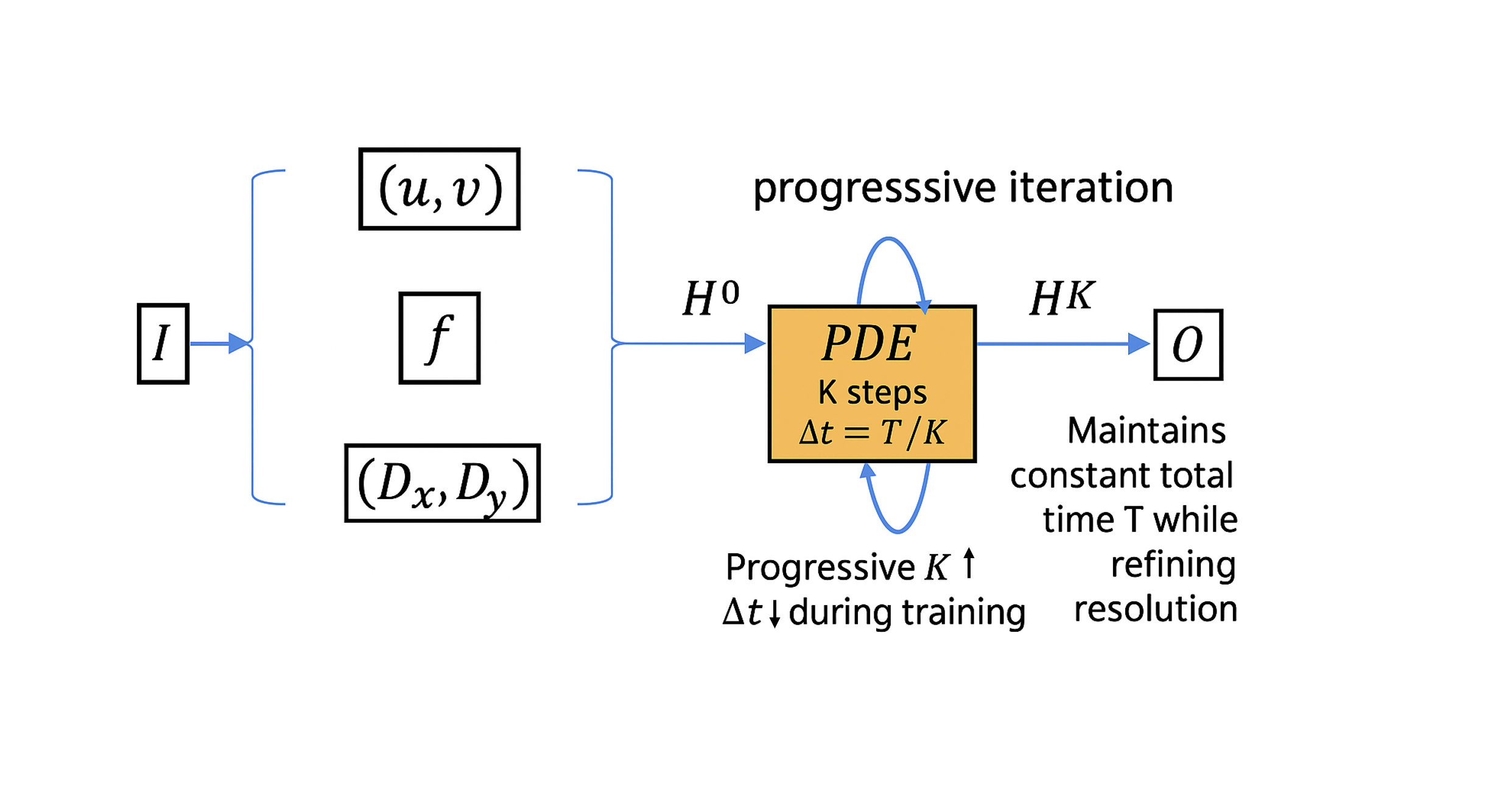}
    \caption{The Global Layer using PDE models motion blur through learned parameters (u, v), blur kernel f, and diffusion coefficients (Dx, Dy). Progressive increase of iteration steps K during training maintains constant total time T while improving computational efficiency, following the diffusion PDE framework of \cite{kag2022}.}
    \label{fig:unet_pde}
\end{figure}

\subsubsection{Progressive Iteration Strategy}
To address these challenges, we employ a progressive iteration strategy where $K$ is gradually increased during training from $K=1$ to $K=5$. To maintain a constant total integration time $T$ for each layer, we correspondingly reduce $\Delta t$ by the same factor such that $T = K \cdot \Delta t = \text{constant}$. This enables the use of finer temporal resolution as the network stabilizes during training.

The strategy proceeds in three phases: Phase 1 begins with $K=1, \Delta t=1.0$ for the first 10 epochs, establishing a stable training baseline. Phase 2 transitions to $K=3, \Delta t=0.333$ for epochs 10--20, introducing moderate temporal refinement. Phase 3 increases to $K=5, \Delta t=0.2$ for the remaining training, achieving fine temporal discretization. This gradual increase mitigates gradient instability while enabling the benefits of multi-iteration PDE solving.

As training progresses and gradient fluctuations diminish, the network can handle finer temporal discretization without encountering the vanishing gradient problems that would occur if starting directly with $K=5$. This approach draws parallels to progressive training methodologies employed in GANs and classical autoencoders, as well as iterative refinement techniques commonly used in numerical optimization. The effectiveness of this progressive schedule is empirically validated in Section~IV-C, where we demonstrate that direct training with $K=5$ fails to converge while progressive training succeeds.

\section{Experiments}

\subsection{Datasets}

To evaluate the effectiveness of our proposed PDE-based global layer with progressive training, we conduct comprehensive experiments on three widely adopted benchmarks for image deblurring:

\begin{itemize}
    \item \textbf{GoPro Dataset}~\cite{nah2017}: A large-scale synthetic blur dataset comprising 3,214 blurred/sharp image pairs extracted from high-frame-rate videos captured with GoPro cameras. The dataset features diverse dynamic scenes with realistic motion blur patterns, making it a standard benchmark for evaluating motion deblurring algorithms under controlled conditions.
    
    \item \textbf{RealBlur-R}~\cite{rim2020}: A real-world blur dataset containing images captured with camera shake using various handheld devices. This dataset provides authentic blur degradation that better reflects practical imaging scenarios, enabling evaluation of our method's robustness to natural blur variations encountered in real photography.
    
    \item \textbf{RealBlur-J}~\cite{rim2020}: A complementary real-world dataset focusing on images with motion blur caused by object movement and camera motion in JPEG format. It offers additional validation for our approach across different types of realistic blur patterns and compression artifacts commonly found in real-world applications.
\end{itemize}

\subsection{Integration into State-of-the-Art Models}

We integrate our PDE-informed global layer into several representative state-of-the-art architectures at the network bottleneck, positioned after the encoder and before the decoder at the lowest spatial resolution:

\begin{itemize}
    \item \textbf{FFTformer}~\cite{kong2023}: A frequency-domain transformer that exploits global dependencies through Fourier transforms for efficient image restoration. We insert the PDE-based global layer after the frequency domain processing blocks, allowing it to refine the globally-aware features before spatial reconstruction.
    
    \item \textbf{NAFNet}~\cite{chen2022}: A baseline architecture designed for efficiency in image restoration with simplified activation-free residual blocks that achieve competitive performance with reduced computational overhead. The PDE layer is integrated within the middle stages of the encoder-decoder structure, specifically after the deepest feature extraction layers where receptive fields are largest.
    
    \item \textbf{Restormer}~\cite{zamir2021}: A transformer-based architecture optimized for image restoration, particularly effective in handling complex degradations through multi-scale feature processing and cross-attention mechanisms. We incorporate the PDE layer between the encoder and decoder sections, enabling global context modeling at the feature bottleneck where spatial resolution is lowest but semantic information is richest.
    
    \item \textbf{Stripformer}~\cite{tseng2022stripformer}: A transformer-based architecture that employs strip-based attention mechanisms for efficient long-range dependency modeling in image restoration tasks. The PDE-based global layer is positioned at the bottleneck of the architecture, at the lowest spatial resolution where it can most effectively capture long-range dependencies and enhance the strip-based attention's ability to model directional blur patterns before the reconstruction phase begins.
\end{itemize}

\subsection{Progressive Training Validation}

To empirically validate that gradient instability prevents direct multi-iteration training, we compare two training strategies on NAFNet with 5 PDE layers on GoPro datasets:

\begin{table}[h]
\centering
\caption{Progressive training is necessary for multi-iteration PDE schemes. Here are the results on the GoPro dataset for the NafNet model.}
\begin{tabular}{|l|c|c|c|}
\hline
Strategy & K & Final PSNR & Converged \\
\hline
Direct & 5 (fixed) & - & (diverged epoch 18) \\
Progressive & 1→3→5 & 34.01 & Yes \\
\hline
\end{tabular}
\end{table}

Direct training with K=5 resulted in loss divergence at epoch 18, with gradient norms exceeding 10³, confirming the gradient instability. Progressive training successfully enabled stable convergence to K=5 by allowing the network to adapt gradually to increasing gradient path lengths.

\subsection{Training Configuration and Progressive Strategy}

\subsubsection{Progressive Training Protocol}
Given the stability constraints identified in Section III-B, we implement a progressive iteration strategy during training of the PDE-enhanced architectures. Each model (FFTformer+PDE, NAFNet+PDE, Restormer+PDE, Stripformer+PDE) is trained end-to-end with the integrated PDE global layers using progressive temporal refinement.

The progressive iteration schedule is implemented as follows:
\begin{itemize}
    \item \textbf{Phase 1} (first 10 epochs): $K=1$, $\Delta t = 1.0$, Total time $T = 1.0$
    \item \textbf{Phase 2} (epochs 10-20): $K=3$, $\Delta t = 0.333$, Total time $T = 1.0$
    \item \textbf{Phase 3} (epoch 20 till end of training): $K=5$, $\Delta t = 0.2$, Total time $T = 1.0$
\end{itemize}

During training, we begin with coarse temporal discretization using larger iteration steps and gradually increase the number of iterations while correspondingly reducing the temporal step size to maintain constant total integration time $T = K \cdot \Delta t$. The PDE parameters (velocity components $u$, $v$, and diffusion coefficients $D_x$, $D_y$) are initialized using Xavier initialization scaled by 0.01. The iteration count $K$ is doubled while the temporal step $\Delta t$ is halved at each phase transition, ensuring consistent total integration time across all phases while enabling progressively finer temporal resolution as the network stabilizes during training.

\subsubsection{Training Hyperparameters}
For all experiments, we use the following training configuration:
\begin{itemize}
    \item \textbf{Training hyperparameters}: Optimizer, learning rate schedule, batch size, training epochs, loss function, and data augmentation are adopted from each architecture's original paper
    \item \textbf{PDE-specific parameters}: 
    \begin{itemize}
        \item Spatial discretization: $\delta x = \delta y = 1.0$
        \item Temporal step: $\delta t = 1.0$ (iteration constraint)
        \item Training epochs: 10 for Phase 1, 10 for Phase 2, same number of iterations as mentioned in the papers for the respective baseline model for Phase 3
    \end{itemize}
\end{itemize}

\begin{table}[t]
\centering
\caption{Quantitative comparison of models with and without the PDE-based global layer on GoPro, RealBlur-R and RealBlur-J datasets. The number of PDE layers for our models is set to 5 and $K=5$.}
\label{tab:results}
\small
\setlength{\tabcolsep}{4pt}
\begin{tabular}{|l|cc|cc|cc|}
\hline
\multirow{2}{*}{Model} & \multicolumn{2}{c|}{GoPro} & \multicolumn{2}{c|}{RealBlur-R} & \multicolumn{2}{c|}{RealBlur-J} \\
 & PSNR & SSIM & PSNR & SSIM & PSNR & SSIM \\
\hline
FFTformer       & 34.21 & 0.969 & 40.11 & 0.973 & 32.62 & 0.933 \\
FFTformer+PDE   & \textbf{34.45} & \textbf{0.971} & \textbf{40.63} & \textbf{0.977} & \textbf{32.92} & \textbf{0.934} \\
\hline
NAFNet & 33.71 & 0.967 & 39.89 & 0.973 & 32.01 & 0.920 \\
NAFNet+PDE      & \textbf{34.01} & \textbf{0.969} & \textbf{40.10} & \textbf{0.974} & \textbf{32.98} & \textbf{0.932} \\
\hline
Restormer       & 32.92 & 0.961 & 40.15 & 0.974 & 32.88 & 0.933 \\
Restormer+PDE & \textbf{33.77} & \textbf{0.968} & \textbf{40.68} & \textbf{0.976} & \textbf{33.15} & \textbf{0.937} \\
\hline
Stripformer     & 33.08 & 0.962 & 39.84 & 0.974 & 32.48 & 0.929 \\
Stripformer+PDE & \textbf{33.79} & \textbf{0.968} & \textbf{40.89} & \textbf{0.977} & \textbf{33.12} & \textbf{0.936} \\
\hline
\end{tabular}
\end{table}

\begin{table}[t]
\centering
\caption{Computational complexity comparison in terms of GMACs for models with and without PDE-based global layer. The number of PDE layers for our models is set to 5 and $K=5$.}
\label{tab:complexity}
\footnotesize
\setlength{\tabcolsep}{15pt}
\begin{tabular}{|l|c|c|c|}
\hline
Model & GMACs & Model & GMACs \\
\hline
FFTformer & 131.45 & FFTformer+PDE & 132.87 \\
NAFNet & 63.64 & NAFNet+PDE & 64.24 \\
Restormer & 141.00 & Restormer+PDE & 141.60 \\
Stripformer & 169.89 & Stripformer+PDE & 172.31 \\
\hline
\end{tabular}
\end{table}
\subsection{Evaluation Protocol}

We conduct a comprehensive evaluation to assess the impact of our PDE-based global layer on restoration performance and computational efficiency. Each baseline architecture is tested both with and without the integrated PDE layer under identical training and testing conditions.

\textbf{Performance Metrics:}
\begin{itemize}
    \item \textbf{PSNR (Peak Signal-to-Noise Ratio)}: Measures pixel-level reconstruction fidelity between restored and ground truth images.
    \item \textbf{SSIM (Structural Similarity Index)}: Evaluates perceptual quality by comparing structural information, luminance, and contrast.
\end{itemize}

\textbf{Computational Efficiency Metrics:}
\begin{itemize}
    \item \textbf{GMACs (Giga Multiply-Accumulate Operations)}: Quantifies computational complexity to assess the overhead introduced by the PDE-based global layers across different architectures.
\end{itemize}

\textbf{Evaluation Setup:} All models are evaluated on the test splits of GoPro, RealBlur-R, and RealBlur-J datasets using input images of size $256 \times 256$ pixels. For computational complexity analysis, we measure GMACs using the final trained models with the complete progressive training protocol ($K=5$, $\Delta t = 0.2$). This comprehensive evaluation framework enables rigorous analysis of how the proposed PDE-informed global layer contributes to restoration quality and the associated computational overhead across multiple backbone architectures.

\begin{table}[t]
\centering
\caption{Performance comparison across different iteration counts $K$ for the NafNet PDE-based global layer models on GoPro, RealBlur-R and RealBlur-J datasets. The number of PDE layers is set to 5.}
\label{tab:iterations}
\scriptsize
\setlength{\tabcolsep}{5pt}
\begin{tabular}{|l|c c|c c|c c|}
\hline
\multirow{2}{*}{Iterations ($K$)} & \multicolumn{2}{c|}{GoPro} & \multicolumn{2}{c|}{RealBlur-R} & \multicolumn{2}{c|}{RealBlur-J} \\
 & PSNR $\uparrow$ & SSIM $\uparrow$ & PSNR $\uparrow$ & SSIM $\uparrow$ & PSNR $\uparrow$ & SSIM $\uparrow$ \\
\hline
$K=0$           & 33.71 & 0.967 & 39.89 & 0.973 & 32.01 & 0.920 \\
$K=1$           & 33.69 & 0.967 & 39.94 & 0.973 & 32.09 & 0.920 \\
$K=3$           & 33.88 & 0.968 & 40.03 & 0.974 & 32.67 & 0.929 \\
$K=5$           & 34.01 & 0.969 & 40.10 & 0.974 & 32.98 & 0.932 \\
$K=7$           & \textbf{34.03} & \textbf{0.969} & \textbf{40.15} & \textbf{0.974} & \textbf{33.00} & \textbf{0.932} \\
\hline
\end{tabular}
\end{table}

\begin{table}[t]
\centering
\caption{Performance comparison across different numbers of PDE-based global layers on NafNet for GoPro, RealBlur-R, and RealBlur-J datasets. $K=5$ for all the results presented in this table.}
\label{tab:layers}
\scriptsize
\setlength{\tabcolsep}{5pt}
\begin{tabular}{|l|c c|c c|c c|}
\hline
\multirow{2}{*}{Number of Layers} & \multicolumn{2}{c|}{GoPro} & \multicolumn{2}{c|}{RealBlur-R} & \multicolumn{2}{c|}{RealBlur-J} \\
 & PSNR $\uparrow$ & SSIM $\uparrow$ & PSNR $\uparrow$ & SSIM $\uparrow$ & PSNR $\uparrow$ & SSIM $\uparrow$ \\
\hline
0 Layer         & 33.71 & 0.967 & 39.89 & 0.973 & 32.01 & 0.920 \\
1 Layer         & 33.80 & 0.967 & 39.94 & 0.972 & 32.15 & 0.923 \\
3 Layers        & 33.92 & 0.968 & 40.02 & 0.973 & 32.57 & 0.927 \\
5 Layers        & \textbf{34.01} & \textbf{0.969} & \textbf{40.10} & \textbf{0.974} & \textbf{32.98} & \textbf{0.932} \\
\hline
\end{tabular}
\end{table}

This setup allows us to rigorously analyze how the proposed PDE-informed global layer contributes to restoration quality, perceptual improvement, and computational efficiency across multiple backbones.

\section{DISCUSSION}

\subsection{Consistent Quality Improvements with Minimal Overhead}

Our approach achieves consistent improvements across all tested architectures and datasets (Table II), with PSNR gains ranging from +0.24 dB to +1.05 dB and SSIM improvements from +0.001 to +0.007. These improvements are particularly notable given the minimal computational overhead, adding only 0.6 to 2.4 GMACs (approximately 1\%) depending on the base architecture.

The universality of improvements across diverse architectures validates the generality of our approach. Both CNN-based (NAFNet) and transformer-based architectures (FFTformer, Restormer, Stripformer) benefit from PDE integration, demonstrating that physics-informed feature evolution provides complementary advantages regardless of the underlying architectural paradigm. This broad applicability suggests that PDE layers can be readily integrated into existing restoration pipelines without architecture-specific modifications.

Importantly, these quality improvements translate to meaningful perceptual benefits. SSIM gains of 0.001-0.007 based on Table \ref{tab:results} indicate enhanced structural preservation and texture fidelity, which are critical for downstream vision tasks and human perception. The combined PSNR and SSIM improvements demonstrate that our method enhances both pixel-level accuracy and perceptual quality—a dual benefit that purely data-driven approaches often struggle to achieve simultaneously.

\subsection{Architecture and Dataset-Specific Performance Insights}

While all architectures benefit from PDE integration, the magnitude of improvement varies meaningfully across base architectures in Table \ref{tab:results}, ranging from +0.38 dB average (FFTformer) to +0.71 dB average (Stripformer). This variance provides valuable insight into deployment strategies. FFTformer's frequency-domain processing already provides strong global receptive fields, so PDE layers offer complementary but smaller gains. Conversely, Stripformer's strip-based attention benefits substantially from the omnidirectional feature propagation provided by PDE layers, suggesting that architectures with directional biases gain most from our approach.

Dataset-specific performance patterns further demonstrate the method's strengths. Based on Table \ref{tab:results} RealBlur-J shows the largest improvements (+0.64 dB average), significantly outpacing gains on GoPro (+0.36 dB average) and RealBlur-R (+0.44 dB average). This pattern reveals a key advantage: PDE-based iterative refinement particularly excels when handling compound degradations. RealBlur-J combines motion blur with JPEG compression artifacts, and the iterative feature evolution appears to disentangle multiple degradation sources more effectively than single-pass convolutions. This suggests our approach is especially valuable for real-world deployment scenarios where images exhibit multiple, interacting degradations.

Tables IV and V reveal optimal configurations that maximize performance while maintaining efficiency. The K iteration ablation shows steady improvements from K=1 to K=5 (+0.30 dB on GoPro), with K=7 providing only marginal additional gains (+0.02 dB). Similarly, the layer count ablation demonstrates that 5 layers achieve near-optimal performance, with diminishing returns beyond this point. These results indicate natural convergence properties rather than arbitrary cutoffs. For deployment, K=3 with 3 layers provides benefits at reduced cost, while K=5 with 5 layers represents the optimal configuration for applications demanding maximum quality. This flexibility enables practitioners to choose appropriate trade-offs for their specific resource constraints and quality requirements.

\section{CONCLUSIONS}

This paper presents a novel approach to motion deblurring by integrating physics-informed partial differential equation (PDE) dynamics into state-of-the-art restoration architectures through a progressive training strategy. We use a PDE-based global layer leveraging advection-diffusion equations to model feature motion deblurring, naturally capturing the directional flow characteristics of motion blur while enabling principled global spatial modeling through physics-informed constraints.

Our comprehensive experiments on standard motion deblurring benchmarks (GoPro, RealBlur-R, and RealBlur-J) validate the effectiveness of physics-informed feature processing. The consistent improvements in PSNR (0.2 to 1.0 dB) and SSIM (0.001 to 0.007) metrics across all tested architectures (FFTformer, NAFNet, Restormer, and Stripformer) based on Table \ref{tab:results} demonstrate that our approach successfully bridges the gap between mathematical modeling and deep learning for image restoration. Importantly, these quality improvements are achieved with minimal computational overhead, adding only approximately 1\% to inference GMACs (0.6 to 2.4 GMACs depending on the base architecture) based on Table \ref{tab:complexity}.

The progressive training strategy addresses critical stability issues inherent in multi-iteration PDE schemes. By gradually increasing the number of PDE iterations from $K=1$ to $K=5$ while correspondingly reducing the temporal step size to maintain constant total integration time, our approach ensures stable gradient flow and convergence throughout the training process. This strategy enables successful integration of iterative PDE solvers into existing restoration architectures without encountering the gradient vanishing problems that would otherwise prevent effective backpropagation.

The results demonstrate that incorporating mathematical physics principles into deep learning architectures can enhance restoration quality while maintaining computational efficiency comparable to baseline methods. This physics-informed approach establishes a promising direction for leveraging domain knowledge through PDE-based modeling in image restoration tasks. By treating blur removal as a guided feature evolution process constrained by advection-diffusion dynamics, our method achieves more effective restoration across diverse blur scenarios.

\balance

\end{document}